\documentclass{article}

\PassOptionsToPackage{numbers, compress}{natbib}

\usepackage[final]{neurips_2019}
\usepackage{wrapfig}
\usepackage{caption}
\usepackage{subcaption}

\PassOptionsToPackage{numbers, compress}{natbib}
\usepackage[table,usenames,dvipsnames,svgnames]{xcolor}

\usepackage{booktabs} %
\usepackage{lineno, blindtext, color}

\usepackage{tcolorbox}
\usepackage{environ}
\usepackage{xargs}                      %

\usepackage{listings}

\definecolor{oldeditorcolor}{rgb}{0.215686,0.494118,0.721569}

\newif\ifsubmit
\submittrue
\ifsubmit
    \usepackage[disable]{todonotes} %
    \newcommand{\usure}[1]{}
    \newcommand{\change}[1]{}
    \newcommand{\info}[1]{}
    \newcommand{\improvement}[1]{}
    \newcommandx{\cheng}[2][1=]{}
    \newcommandx{\abdul}[2][1=]{}
    \newcommandx{\abi}[2][1=]{}
    \newcommandx{\renee}[2][1=]{}
    \newcommandx{\junjun}[2][1=]{}
    \newcommandx{\wenmei}[2][1=]{}
    \newcommandx{\old}[2][1=]{}
    \newcommandx{\delete}[2][1=]{}
\else
\usepackage{comment}
\usepackage[
    colorinlistoftodos,
    prependcaption,
    draft,
    textsize=tiny,
    obeyFinal,
    textwidth=2cm
]{todonotes}
	 
	\newcounter{adtodocounter} 
	\newcounter{cltodocounter} 
	\newcounter{jjtodocounter} 
	\newcounter{wmhtodocounter} 
	\newcounter{abtodocounter} 
	\newcounter{rntodocounter} 
	\newcounter{oldtodocounter} 
	\newcounter{deletetodocounter} 
    \newcommandx{\unsure}[2][1=]{\todo[linecolor=red,backgroundcolor=red!25,bordercolor=red,#1]{#2}}
    \newcommandx{\change}[2][1=]{\todo[linecolor=blue,backgroundcolor=blue!25,bordercolor=blue,#1]{#2}}
    \newcommandx{\info}[2][1=]{\todo[linecolor=OliveGreen,backgroundcolor=OliveGreen!25,bordercolor=OliveGreen,#1]{#2}}
    \newcommandx{\improvement}[2][1=]{ \marginpar[\todo[linecolor=Plum,backgroundcolor=Plum!25,bordercolor=Plum,#1]{#2}]{}}
    \newcommandx{\thiswillnotshow}[2][1=]{\todo[disable,#1]{#2}}
    \newcommandx{\abi}[2][1=]{\stepcounter{abtodocounter} \todo[linecolor=red,backgroundcolor=purple!25,bordercolor=purple,#1]{AB(\thecltodocounter): #2}}
    \newcommandx{\abdul}[2][1=]{\stepcounter{adtodocounter} \todo[linecolor=red,backgroundcolor=red!25,bordercolor=red,#1]{AD(\theadtodocounter): #2}}
    \newcommandx{\cheng}[2][1=]{\stepcounter{cltodocounter} \todo[linecolor=teal,backgroundcolor=teal!25,bordercolor=teal,#1]{CL(\theabtodocounter): #2}}
    \newcommandx{\renee}[2][1=]{\stepcounter{rntodocounter} \todo[linecolor=yellow,backgroundcolor=yello!25,bordercolor=OliveGreen,#1]{RN(\therntodocounter): #2}}
    \newcommandx{\jinjun}[2][1=]{\stepcounter{jjtodocounter} \todo[linecolor=yellow,backgroundcolor=yellow!25,bordercolor=yellow,#1]{JJ(\thejjtodocounter): #2}}
    \newcommandx{\wenmei}[2][1=]{\stepcounter{wmhtodocounter} \todo[linecolor=Plum,backgroundcolor=Plum!25,bordercolor=Plum,#1]{WH(\thewmhtodocounter): #2}}
    \newcommandx{\old}[2][1=]{\stepcounter{oldtodocounter} \todo[linecolor=oldeditorcolor,backgroundcolor=oldeditorcolor!25,bordercolor=oldeditorcolor,#1]{OLD(\theoldtodocounter): #2}}
    \newcommandx{\delete}[2][1=]{\stepcounter{deletetodocounter} \todo[linecolor=oldeditorcolor,backgroundcolor=oldeditorcolor!25,bordercolor=oldeditorcolor,#1]{OLD(\thedeletetodocounter): #2}}
\fi

\definecolor{yamlred}{rgb}{0.843137,0.188235,0.152941}
\definecolor{yamlblue}{rgb}{0.27451,0.32549,0.384314}
\definecolor{yamlkey}{rgb}{0.980392,0.501961,0.447059}

\newcommand\YAMLfontsize{\fontsize{6}{6}}
\newcommand\YAMLfontstyle{\ttfamily}
\newcommand\YAMLcolonstyle{\color{black}\mdseries\YAMLfontsize\YAMLfontstyle}
\newcommand\YAMLkeystyle{\bfseries\color{yamlblue}\YAMLfontsize\YAMLfontstyle}
\newcommand\YAMLvaluestyle{\color{black}\mdseries\YAMLfontsize\YAMLfontstyle}

\definecolor{yamlgray}{rgb}{0.5,0.5,0.5}
\lstdefinelanguage{yaml}
{
  keywords={true,false,null,y,n},
  keywordstyle=\YAMLvaluestyle,
  basicstyle=\YAMLfontsize\YAMLkeystyle,                                 %
  sensitive=false,
  comment=[l]{\#},
  morecomment=[s]{/*}{*/},
  commentstyle=\YAMLcolonstyle\color{yamlred}\YAMLfontstyle\YAMLfontsize\YAMLfontstyle,
  stringstyle=\YAMLvaluestyle\YAMLfontstyle\YAMLfontsize\YAMLfontstyle,
  moredelim=[l][\color{orange}]{\&},
  moredelim=[l][\color{yamlred}]{*},
  moredelim=**[il][\YAMLcolonstyle{:}\YAMLvaluestyle]{:},   %
  morestring=[b]',
  morestring=[b]",
  literate =    {---}{{\ProcessThreeDashes}}3
                {>}{{\textcolor{yamlred}\textbar}}1     
                {|}{{\textcolor{yamlred}\textbar}}1 
                {\ -\ }{{\mdseries\ -\ }}3,       
    frame=top,
    frame=bottom,
    numbers=left,
    rulesep=1pt,
	breaklines=true,
    mathescape=true,
    xleftmargin=2em,
    framexleftmargin=3em,
    stepnumber=1,
    escapechar=|,
    captionpos=t,
    numberstyle=\tiny\color{yamlgray},
}

\makeatother

\newcommand\ProcessThreeDashes{\llap{\color{cyan}\mdseries-{-}-}}

\usepackage[hidelinks]{hyperref}
\hypersetup{draft}

\usepackage{latexsym}
\usepackage{verbatim}
\usepackage{graphicx}
\usepackage{subcaption} 
\usepackage[font={small}]{caption}
\usepackage{booktabs} %
\usepackage{soul}
\usepackage{amsthm}
\usepackage{amsmath}
\usepackage[utf8]{inputenc}
\usepackage{xspace}
\usepackage{array}
\usepackage{makecell}
\usepackage{rotating}
\usepackage{filecontents}
\usepackage{pgfplots}
\usepackage{pgfplotstable}
\usepackage{tikz}
\usepackage[normalem]{ulem}
\usepackage{enumitem}
\usepackage{amssymb}%

\usepackage{times}
\usepackage{url}
\makeatletter
\newsavebox{\measure@tikzpicture}
\NewEnviron{scaletikzpicturetowidth}[1]{%
  \def\tikz@width{#1}%
  \begin{lrbox}{\measure@tikzpicture}%
  \BODY
  \end{lrbox}%
  \pgfmathparse{#1/\wd\measure@tikzpicture}%
  \BODY
}

\setlist{noitemsep,nolistsep}

\newcommand{\cmmnt}[1]{\ignorespaces}

\definecolor{myred}{rgb}{0.843137,0.188235,0.152941}
\definecolor{myblack}{rgb}{0.27451,0.32549,0.384314}
\definecolor{mygreen}{rgb}{0.301961,0.686275,0.290196}
\definecolor{myyellow}{rgb}{0.996078,0.878431,0.564706}
\definecolor{myblue}{rgb}{0.568627,0.74902,0.858824}

\pgfplotsset{compat=newest,}
\usepgfplotslibrary{external}
\usetikzlibrary{babel}

\pgfplotsset{every axis/.style={scale only axis}}

\definecolor{plotcolor1}{rgb}{0.568627,0.74902,0.858824}
\definecolor{plotcolor2}{rgb}{0.996078,0.878431,0.564706}
\definecolor{plotcolor3}{rgb}{0.27451,0.32549,0.384314}
\definecolor{plotcolor4}{rgb}{0.843137,0.188235,0.152941}
\definecolor{plotcolor5}{rgb}{0.988235,0.552941,0.34902}
\definecolor{plotcolor6}{rgb}{0.596078,0.305882,0.639216}
\definecolor{plotcolor7}{rgb}{0.65098,0.337255,0.156863}
\definecolor{plotcolor8}{rgb}{0.105882,0.619608,0.466667}
\definecolor{plotcolor9}{rgb}{1.,1.,0.6}
\definecolor{plotcolor10}{rgb}{0.745098,0.729412,0.854902}
\definecolor{plotcolor11}{rgb}{0.984314,0.501961,0.447059}
\definecolor{plotcolor12}{rgb}{0.501961,0.694118,0.827451}
\definecolor{plotcolor13}{rgb}{1.,1.,0.2}
\definecolor{plotcolor14}{rgb}{0.992157,0.705882,0.384314}
\definecolor{plotcolor15}{rgb}{0.988235,0.803922,0.898039}
\definecolor{plotcolor16}{rgb}{0.701961,0.870588,0.411765}
\definecolor{plotcolor17}{rgb}{0.215686,0.494118,0.721569}
\definecolor{plotcolor18}{rgb}{0.941176,0.231373,0.12549}
\definecolor{plotcolor19}{rgb}{0.168627,0.54902,0.745098}
\definecolor{plotcolor20}{rgb}{0.552941,0.827451,0.780392}
\definecolor{plotcolor21}{rgb}{0.968627,0.505882,0.74902}
\definecolor{plotcolor22}{rgb}{0.6,0.6,0.6}
\definecolor{plotcolor23}{rgb}{0.301961,0.686275,0.290196}
\definecolor{plotcolor24}{rgb}{0.980392,0.501961,0.447059}

\usepgfplotslibrary{colorbrewer}
\pgfplotsset{cycle list/Dark2-8}

\newcommand{\carml}{MLModelScope\xspace}

\newcommand{\ignore}[1]{}

\DeclareRobustCommand*\circled[1]{\tikz[baseline=(char.base)]{
            \node[shape=circle,white,draw,fill=black,inner sep=0.0pt] (char) {\small #1};}}

\newcommand*\circledwhite[1]{\tikz[baseline=(char.base)]{
            \node[shape=circle,draw,inner sep=0.0pt] (char) {\small  #1};}}

\newcommand{\floor}[1]{\left\lfloor #1 \right\rfloor}

\newcolumntype{C}{>{\centering\arraybackslash} m{.1\linewidth} }  %

\usepackage{times}
\usepackage{epsfig}
\usepackage{graphicx}
\usepackage{amsmath}
\usepackage{amssymb}
\usepackage{enumitem}
\usepackage{enumitem}
\usepackage{kantlipsum}

\usepackage[utf8]{inputenc} %
\usepackage[T1]{fontenc}    %
\usepackage{hyperref}       %
\usepackage{url}            %
\usepackage{booktabs}       %
\usepackage{amsfonts}       %
\usepackage{nicefrac}       %
\usepackage{microtype}      %

\usepackage{xspace}
\usepackage{filecontents}
\usepackage{pgfplots}
\usepackage{pgfplotstable}
\usepackage{multirow}
\usepackage{tikz}
\usepackage[normalem]{ulem}
\usepackage{enumitem}
\usepackage{listings}
\usepackage{tcolorbox}
\usepackage{xargs}                      %
\usepackage{amssymb}%
\usepackage{pifont}%

\usepackage{environ}%

\tcbuselibrary{listings,breakable}

\definecolor{myred}{rgb}{0.843137,0.188235,0.152941}
\definecolor{myblack}{rgb}{0.27451,0.32549,0.384314}
\definecolor{mygreen}{rgb}{0.301961,0.686275,0.290196}
\definecolor{myyellow}{rgb}{0.996078,0.878431,0.564706}
\definecolor{myblue}{rgb}{0.568627,0.74902,0.858824}

\input{glyphtounicode}
\title{Frustrated with Replicating Claims of a Shared Model? A Solution}

\begin{document}

\author{Abdul Dakkak \textsuperscript{*} \\
   University of Illinois, Urbana-Champaign \\
    Champaign, IL 61820 \\
    \texttt{dakkak@illinois.edu}
    \And
    Cheng Li \textsuperscript{*} \\
   University of Illinois, Urbana-Champaign \\
    Champaign, IL 61820 \\
    \texttt{cli99@illinois.edu} 
    \And
    Jinjun Xiong \\
    IBM Thomas J. Watson Research Center \\
    Yorktown Heights, NY 10598 \\
    \texttt{jinjun@us.ibm.com}
    \And
    Wen-mei Hwu \\
   University of Illinois, Urbana-Champaign \\
    Champaign, IL 61820 \\
    \texttt{w-hwu@illinois.edu}
}

\renewcommand{\thefootnote}{\fnsymbol{footnote}}
\footnotetext[1]{The two authors contributed equally to this paper.}

\date{}

\maketitle

\begin{abstract}

Machine Learning (ML) and Deep Learning (DL) innovations are being introduced at such a rapid pace that model owners and evaluators are hard-pressed analyzing and studying them.
This is exacerbated by the complicated procedures for evaluation.
The lack of standard systems and efficient techniques for specifying and provisioning ML/DL evaluation is the main cause of this ``pain point”.
This work discusses common pitfalls for replicating DL model evaluation, and shows that these  subtle pitfalls can affect both accuracy and performance.
It then proposes a solution to remedy these pitfalls called \carml, a specification for repeatable model evaluation and a runtime to provision and measure experiments.
We show that by easing the model specification and evaluation process, \carml facilitates rapid adoption of ML/DL innovations.

\end{abstract}

\section{Introduction}\label{sec:intro}

The recent impressive progress made by Machine Learning (ML) and Deep Learning (DL) in a wide array of applications, as evident in the number of academic publications, has led to increased public interests in learning and understanding these ML and DL innovations.
Helping the public (i.e., \textit{model evaluators}) understand how various ML/DL models operate is important for such a new technology to be widely adopted in solving problems directly connected to their daily lives, such as autonomous vehicles, face recognition, fraud detection, and loan application.
To facilitate this process, ML/DL model researchers (i.e., \textit{model owners}) are equally eager to explain their models by writing academic papers and sharing their models by open sourcing their code and model parameters.
This has resulted in a flood of ML/DL related \textit{arXiv} papers, conference submissions, GitHub projects, and various ML/DL model repositories. 
These practices have also helped many people in academia to learn from each others' work, leverage each others' models, and propose new ML/DL designs,
which generates a positive feedback cycle, spurring further innovations. 

\abdul{One result of this is that many innovations are made anthropomorphic because people cannot test them out}
These current practices, however, have provided little help to the general non-expert public in understanding or leveraging these ML/DL models.
Even for the the academic experts, understanding and replicating others' openly shared models has been achieved, typically, with heroic efforts. 
In fact, anyone who has attempted to reproduce the results from a paper or open source code can attest how painful the experience is.
People who wish to use a model typically have to download the source code and the model, read the documentation (if any, with usually missing key information) provided by the model owners, configure the hardware/software environment conforming to the requirement (and likely these configurations differ from or, worse, conflict with their existing ones), and, at the end, they may still be unable to reproduce the results reported by the  model owners in the published work.

The issues may or may not be solved by e-mailing the model owners with questions and (hopelessly) waiting for an answer. Or, more likely, one
has to take it upon oneself to investigate the issues by reading the source code.
This phenomenon can be easily confirmed (or become evident) by the many publications and blogs complaining that they cannot reproduce others'
model results for comparison purposes.
It can also be confirmed by counting how many issues on Github exist for those popular papers with
open-source models.
If an ML/DL model expert needs to jump so many hoops to understand an openly shared model, how can we expect these practices to help the general public in gaining such an understanding comfortably?

Simply pointing the blame at model owners for their seemingly lousy job of sharing their model is not fair.
Writing a well-specified document for people to repeat a model is not trivial and takes a great amount of time, which most researchers are lack of.
Even if model owners would spend the time, the complexity and interaction among ML/DL applications, models, frameworks, software stacks, system libraries, and hardware configurations (as shown in this paper) often make writing an effective document a daunting endeavor.
Therefore, there is an urging need to provide a scalable and easy-to-use solution to this problem within the ML/DL community.

In this paper, we first discuss why model sharing by itself is not enough to address this problem and of the main sources of complexity.
We then present a number of pitfalls, some may be seemingly obvious and some are not obvious at all.
We show how both model owners and model evaluators encounter these pitfalls and detail how we can alleviate the issues encountered.
Based on this analysis, we then propose an elegant solution to address this problem.
Our solution is \carml with a number of key contributions:
\circledwhite{1} A text based specification that standardizes model sharing while avoiding the identified pitfalls. With such a specification, model owners can easily share their models without 
writing a full documentation.
\circledwhite{2} A runtime system to provision and manage model evaluation. This runtime system uses model specifications as input and makes model evaluation simple and accessible for general public and experts alike.
\circledwhite{3} A scalable data collection and analysis pipeline that helps ease model understanding, analysis, and comparison.
We discuss the design of \carml in detail,  and how we make such a system scalable, extensible, and easy to use.

Although our system design is general, for illustration purpose, we focus on DL models (such as image classification, object detection etc.).
We implemented \carml as a framework and hardware agnostic evaluation platform, with current support for 
Caffe, Caffe2, CNTK, MXNet, PyTorch, TensorFlow, TensorRT, and TFLite,
running on ARM, Power, and x86 with CPU, GPU, and FPGA.
\carml is not restricted to DL and can be generalized to other application areas where evaluating and replicating claims is also a challenge.

\begin{wrapfigure}{hr}{0.43\textwidth}
  \vspace{-60pt}
  \begin{center}
  \includegraphics[clip, width=0.43\textwidth]{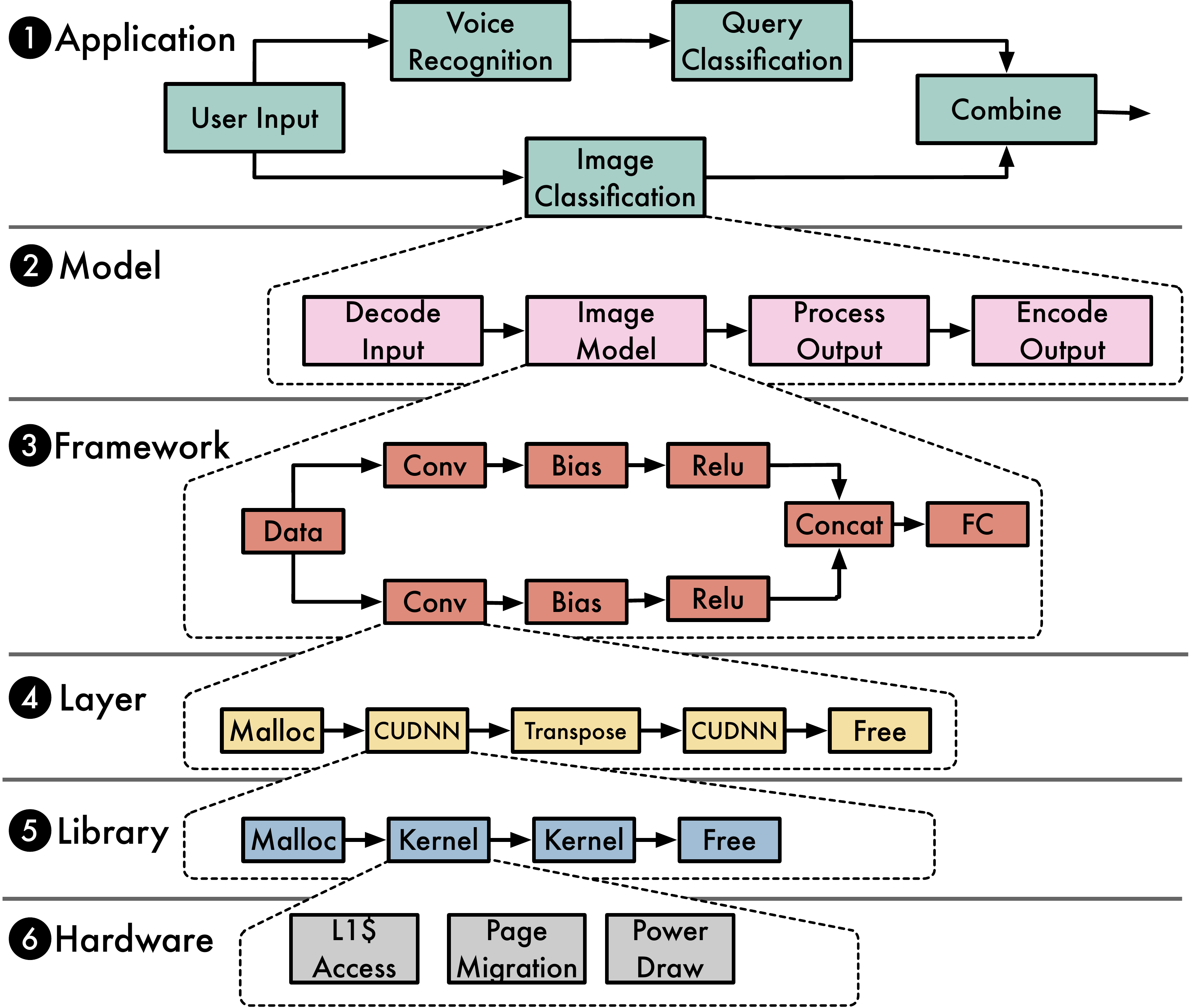}
  \caption{Execution of an AI application at different levels of hardware and software abstractions.
  }
  \label{fig:profile_levels}
  \end{center}
  \vspace{-45pt}
\end{wrapfigure}

\section{ML/DL Model Evaluation Challenges}\label{sec:complexity}

Before we propose a solution, let us first understand the challenges of model evaluation and describe some of the pitfalls a model evaluator might encounter while attempting to replicate a model's claims.

\subsection{Hardware and Software Stack Complexity}

Replicating model evaluation is complex.
Researchers that publish and share ML models can attest to that, but are sometimes  unaware of the full scope of this complexity.
Model evaluation leverages multiple HW/SW abstractions.
All the HW/SW abstractions must work in unison to maintain reported claims of accuracy, performance, and efficiency.
When things go awry, each level within the abstraction hierarchy can be the source of error.
These abstraction levels (shown in 
Figure~\ref{fig:profile_levels}) include:
  \circled{1} \textbf{Application} pipelines which may span multiple machines and employ a suite of models and frameworks.
  \circled{2} \textbf{Model} which defines processing transformations for input and output handling.
  \circled{3} \textbf{Framework} which runs the model by executing the network-layer graph.
  \circled{4} \textbf{Layer} that execute a sequence of library calls.
  \circled{5} \textbf{Library} that invokes a chain of system runtime functions.
Last but not the least, is the \circled{6} \textbf{hardware} which executes the hardware instructions, reads from disks, communicates via networks, and performs other low-level system device operations.
Specifying and setting up the HW/SW stack properly is not easy, and
it is almost an unreasonable ask for model owners
to write a detailed documentation to specify
all the required pieces. 
A solution is needed to address these challenges and simplify the provisioning of the stack.

\subsection{Pitfalls of Replicating Model Claims}\label{sec:spec}

In the interest of space, we take vision models as a proxy, and highlight four key pitfalls that model evaluators may encounter when running a model to reproduce its claims.

\textbf{Pre/Post-Processing}: Pre-processing transforms input into a form that can be consumed by the model whereas post-processing transforms the model output to a form that can be evaluated using metrics or consumed by subsequent components in the application pipeline.
This seemingly lackluster process is, however,
surprisingly subtle and can easily
introduce discrepancies in the results. Some of discrepancies
may be 
``silent errors'' where the evaluation is fine for the majority of the inputs, but  are incorrect for a small number of cases. 
These errors are difficult to identify and even more difficult to debug. Some of these errors will be shown in Section~\ref{sec:pitfalls}. Therefore, an effective solution
needs to allow clear specification of processing parameters and the order of processing.

\textbf{Software Stack}: The major software components for model evaluation are frameworks (e.g. TensorFlow, MXNet, and PyTorch) and libraries (e.g. MKL-DNN, OpenBLAS, and cuDNN).
As our results in Section~\ref{sec:pitfalls} show, different libraries, though functionally
the same, may affect the evaluation's accuracy or performance.
An effective solution needs to help fully specify versions of the framework and libraries and even their compilation options.

\textbf{Hardware Configuration}: Different hardware configurations for diverse architectures can result in varying performance and accuracy.
These configurations include CPU Scaling, multi-threading and vectorization.
It is worthwhile to note that these hardware configurations, cannot be specified within containers. So quite contrary to common perception,  a container based solution is not sufficient.

\begin{wrapfigure}{r}{0.45\textwidth}
  \vspace{-15pt}
  \centering%
\begin{lstlisting}[
    language=yaml,
    escapeinside={(*}{*)},
    floatplacement=H,
    escapechar=|
]
|\label{line:start_model_id}|name: Inception-v3 # model name
version: 1.0.0 # semantic version of model 
|\label{line:end_model_id}|task: classification # model modality
licence: MIT # model licence
description: ...
|\label{line:start_framework_id}|framework: # framework information
  name: TensorFlow
|\label{line:end_framework_id}|  version: ^1.x # framework version constraint
|\label{line:start_container}|container: # available containers 
  amd64:
    cpu: mlcn/tensorflow:1-13-0_amd64-cpu
    gpu: mlcn/tensorflow:1-13-0_amd64-gpu
  ppc64le:
    cpu: mlcn/tensorflow:1-13-0_ppc64le-cpu
|\label{line:end_container}|    gpu: mlcn/tensorflow:1-13-0_ppc64le-gpu
|\label{line:start_env}|envvars:
|\label{line:end_env}|  - TF_ENABLE_WINOGRAD_NONFUSED: 0
|\label{line:start_inputs}|inputs: # model inputs
  - type: image  # first input modality
    layer_name: data
    element_type: float32
    processing: # pre-processing steps
      decode:
        element_type: int8
        data_layout: NHWC
        color_layout: RGB
      crop:
        method: center
        percentage: 87.5
      resize:
        dimensions: [3, 299, 299]
        method: bilinear
        keep_aspect_ratio: true
      mean: [127.5, 127.5, 127.5]
|\label{line:end_inputs}|      rescale: 127.5
|\label{line:start_outputs}|outputs: # model outputs
  - type: probability # output modality
    layer_name: prob
    element_type: float32
    processing:
|\label{line:end_outputs}|      features_url: https://.../synset.txt 
|\label{line:start_weights}|source: # model source 
|\label{line:end_weights}|  graph_path: https://.../inception_v3.pb
|\label{line:start_attrs}|training_dataset:  # dataset used for training
  name: ILSVRC 2012
|\label{line:end_attrs}|  version: 1.0.0 
\end{lstlisting}
\vspace{-.15in}
\captionof{lstlisting}{\carml's model manifest for Inception-v3.
}
\vspace{-40pt}
\label{lst:model_manifest}
\end{wrapfigure}

\textbf{Programming Language}. There is a myth that modern programming languages have a minimum performance impact due to good compilers and faster CPUs.
The choice of programming languages has thus been mostly made based on convenience.
This in part explains the dominance of Python in ML communities.
As our results show, not only is the difference between Python and C/C++  can be significant, but the choice of numeric representation within the same Python language can be significant too.
So a solution must minimize the performance impact due to programming languages for different frameworks used by model evaluators.

\section{\carml Design}\label{sec:design}

\begin{wrapfigure}{tr}{0.43\textwidth}
  \vspace{-50pt}
  \centering%
  \includegraphics[width=0.43\textwidth]{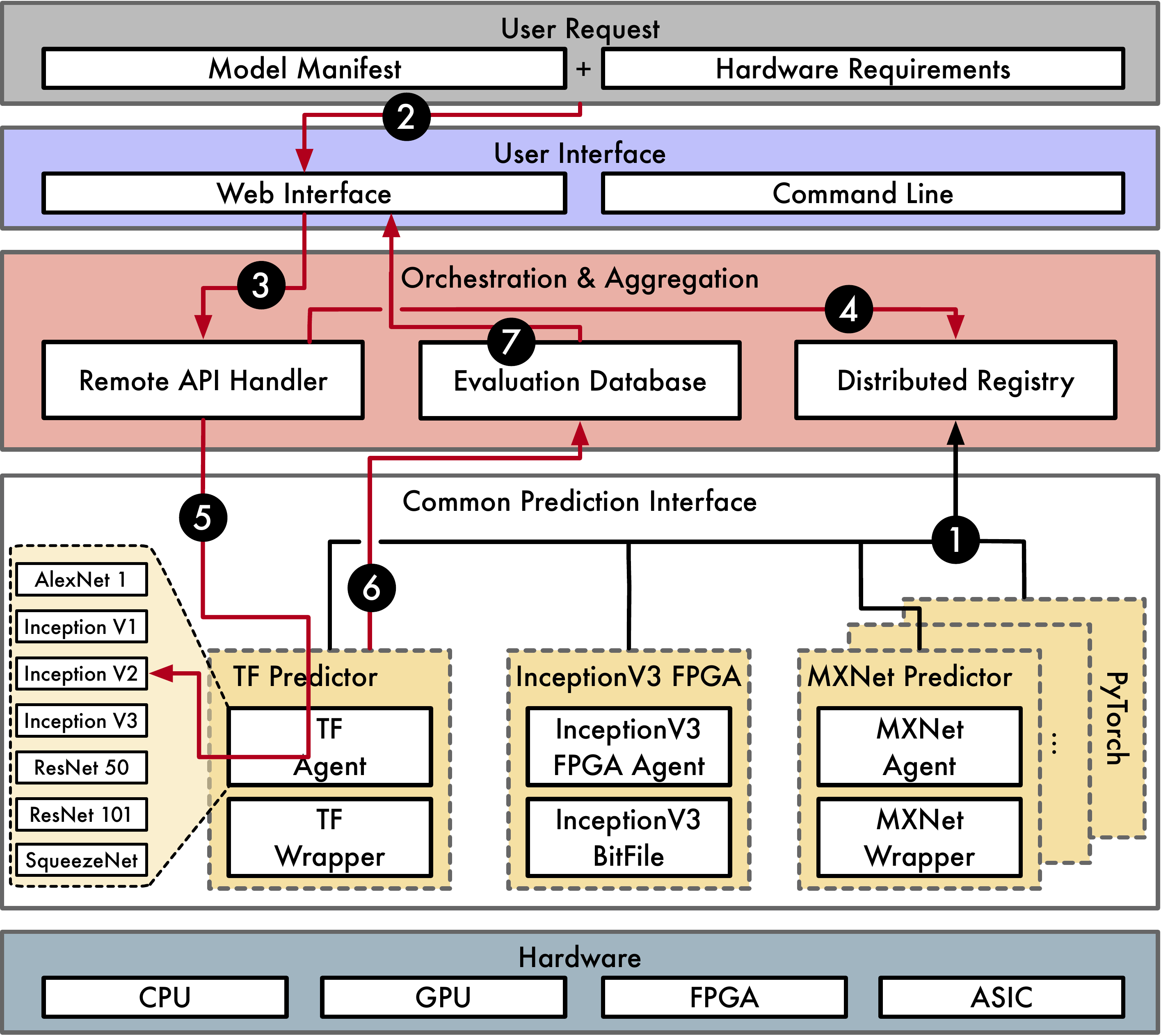}
  \caption{
  \carml's runtime pipeline enables easy, scalable, and repeatable model evaluation across frameworks, models, and systems.
  }
  \label{fig:arch}
 \vspace{-15pt}
\end{wrapfigure}

To address the challenges outlined in Section~\ref{sec:complexity}, we propose \carml, an 
effective system solution for specifying and running model evaluation.
\carml addresses the challenges and pitfalls with a model specification (referred to as model manifest), a distributed runtime to set up the required environments for running model evaluations, and an aggregation and summarization pipeline that captures application traces and explains the model execution process.
This section describes the key components of \carml's design (shown in Figure~\ref{fig:arch}).

\subsection{Model Manifest}\label{sec:manifest}

To avoid the pitfalls mentioned in Section~\ref{sec:spec}, \carml specifies the model evaluation specification via a model manifest and user-provided hardware configuration options.
The manifest is carefully designed to capture information such as software stack used for model evaluation along with other metadata used for experiment management.
The hardware details are not present in the manifest file, but are user options when performing the evaluation, thus allowing a manifest to be evaluated across hardware.
Listing~\ref{lst:model_manifest} shows an example manifest file with key elements as follows:

\textbf{Software Stack} --- \carml uses containers to maintain  software stacks for different system architectures.
Users can select the software stack explicitly or can choose software versions constraints (see below).
\carml not only allows people to build their own containers to be used with the platform, but also provides ready-made containers for all popular frameworks.

\textbf{Inputs and Outputs and Pre-/Post-Processing} --- A model can have multiple inputs and outputs, all dependent on the model's modality.
For image models, \carml is able to perform pre-/post-processing operations for these inputs and outputs without writing code, e.g. image decoding, resizing, and normalization for image input.
As shown in our experiments, the order of operations can have a significant impact on model accuracy. Hence we explicitly require the specification of the order of pre/post-processing operations in the manifest.

\textbf{Versioning} --- 
Models, datasets, and frameworks are all versioned within \carml using a semantic versioning~\cite{semanticver} scheme.
Users request an evaluation by specifying model, framework, and dataset  version constraints along with the target hardware requirements.
\carml uses these constraints to query previous evaluations or schedule a new execution.

All models in \carml are described using model manifest files.
The manifest for Inception-v3 is shown in Listing~\ref{lst:model_manifest} and contains model name, version, and type of task (Lines~\ref{line:start_model_id}--\ref{line:end_model_id}), framework name and version constraint (Lines~\ref{line:start_framework_id}--\ref{line:end_framework_id}),  containers used for evaluation (Lines~\ref{line:start_container}--\ref{line:end_container}), model inputs and  pre-processing steps (Lines~\ref{line:start_inputs}--\ref{line:end_inputs}), model outputs and  post-processing steps (Lines~\ref{line:start_outputs}--\ref{line:end_outputs}), model resources (Lines~\ref{line:start_weights}--\ref{line:end_weights}), and other attributes (Lines~\ref{line:start_attrs}--\ref{line:end_attrs}).
Both the pre- and post-processing steps are executed in the order presented in the manifest file.
E.g. in this manifest, cropping is performed before resizing and normalization.

\subsection{\carml's Runtime Design}

The \carml runtime is designed 
to enable easy, scalable, and repeatable model evaluation.
The key components of \carml's runtime architecture are as follows:

\textbf{Framework and Model Predictors} ---
At the core of the software stack are the ML/DL frameworks.
To enable uniform evaluation and maximize code reuse, \carml abstracts each framework evaluation's C++ API into a consistent interface (called wrapper).
Common code around the wrapper (called \textit{predictor agent}) performs container launching, manifest file handling, downloading of required assets, collecting of performance profiles, and publishing of results.
These agents are distributed, but are managed by a central server.
\carml does not require modifications to a framework, allowing both pre-compiled binary versions of the frameworks (e.g. distributed through Python's pip) and customized versions of a framework.

Some hardware, such as FPGAs or ASICs, cannot run containers and do not have a framework per se.
Since the software stack for these hardware is restrictive (or non-existent), maintaining the software stack is less crucial than for general purpose hardware.
With respect to \carml, both FPGAs and ASICs are exposed as a custom predictor agent.
In the context of FPGA, the \carml predictor agent is a program that is able to download a bitfile (specified in the manifest) and load it onto the FPGA before performing inference.

\textbf{Profilers and Tracers} ---
To enable performance debugging and understanding of application pipelines, \carml collects system, framework, and application level profiling information.
This data is published into a tracing server~\cite{opentracing} where it gets aggregated and summarized.
Through the trace, users can get a holistic view of the application components and identify bottlenecks.
To minimize profiling overhead, the profilers are only enabled when a user triggers them.

\textbf{Web, Command Line, and Library User Interface} ---
We design \carml so that it can facilitate broad usage, 
either as an application or a standalone library to be integrated within existing C/C++, Python, or Java applications.
When used as an application, a user interacts with \carml through its website or command line.
The website and command line interfaces allow users to evaluate models without a lengthy setup of the HW/SW stack.
Users who wish to integrate \carml within their existing tools or pipelines can use \carml's REST or RPC APIs, and interact directly with the remote API handler component.

\subsection{\carml Evaluation Flow}\label{sec:flow}

To illustrate the execution flow, 
consider a user wanting  to evaluate Inception-v3 which was trained using ILSVRC 2012   and run it on an Intel system with an NVLink interconnect and TensorFlow satisfying the \texttt{"$\geq$1.10.x and $\leq$1.13.0"} version constraint. 
The user can specify such requirements using \carml's web UI.
Figure~\ref{fig:arch} shows the evaluation flow of the user's request.
\circled{1}  On system startup, each predictor agent publishes the hardware it is running on, along with the frameworks and models it supports. 
The information is published to a registry, and is made visible to the rest of the \carml system.
\circled{2} A user can then use the web UI to request an evaluation by specifying the model, framework, and hardware constraints.
\circled{3} The web interface performs an API request to the remote API handler, which then \circled{4} queries the registry to find a predictor agent which satisfies the software and hardware constraints provided.
\circled{5} The constraints are then forwarded to one (or all) of the predictors capable of running the model.
The predictor then provisions the hardware and software environment and runs the model.
\circled{6} The predictor then collects and publishes the outputs as well as user-specified profiling information to a centralized evaluation database.
\circled{7} Finally, the web UI generates summary plots and tables of the evaluation results.

\carml's careful design of the specification, runtime, and evaluation flow, reduces time-to-test for both model owners and evaluators.
The design enables a uniform evaluation methodology and better sharing of models and evaluation results.

\section{Experimentation}\label{sec:exper}

We implemented the \carml design presented in Section~\ref{sec:design} and tested using popular hardware stacks (X86, PowerPC, and ARM CPUs as well as GPU and FPGA accelerators) and frameworks (Caffe, Caffe2, CNTK, MXNet, PyTorch, TensorFlow, TensorRT, and TFLite).
We then populated it with over $300$ models (with a wide array of inference tasks) and with the corresponding validation datasets.
For illustration purposes, however, we only present \carml in the context of  image classification.

\begin{figure}[t]
    \centering
\begin{minipage}[t]{0.3\linewidth}
    \centering
    \includegraphics[width=.8\textwidth]{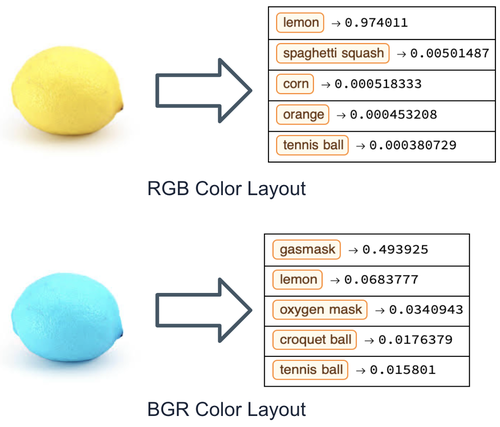}
    \caption{Top 5 predictions using Inception-v3 with RGB or BGR color layout.}
    \label{fig:lemon}
\end{minipage}%
    \hfill%
\begin{minipage}[t]{0.3\linewidth}
\centering
\includegraphics[width=1.0\textwidth]{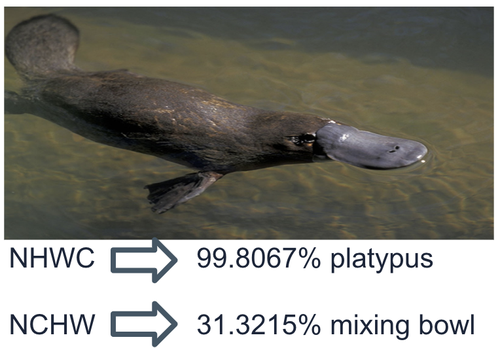}
\caption{Top 1 predictions using Inception-v3 with \texttt{NCHW} or \texttt{NHWC} data layout.}
\label{fig:platapus}
\end{minipage} %
    \hfill%
\begin{minipage}[t]{0.35\linewidth}
\centering
\includegraphics[width=1.0\textwidth]{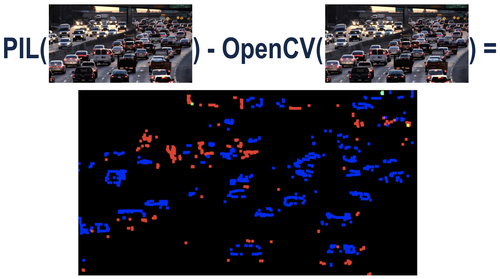}
\caption{Image decoding difference between PIL and OpenCV.}
\label{fig:pil_vs_opencv_diff}
\end{minipage} 

\begin{minipage}[t]{\linewidth}
\centering
\includegraphics[width=0.95\textwidth]{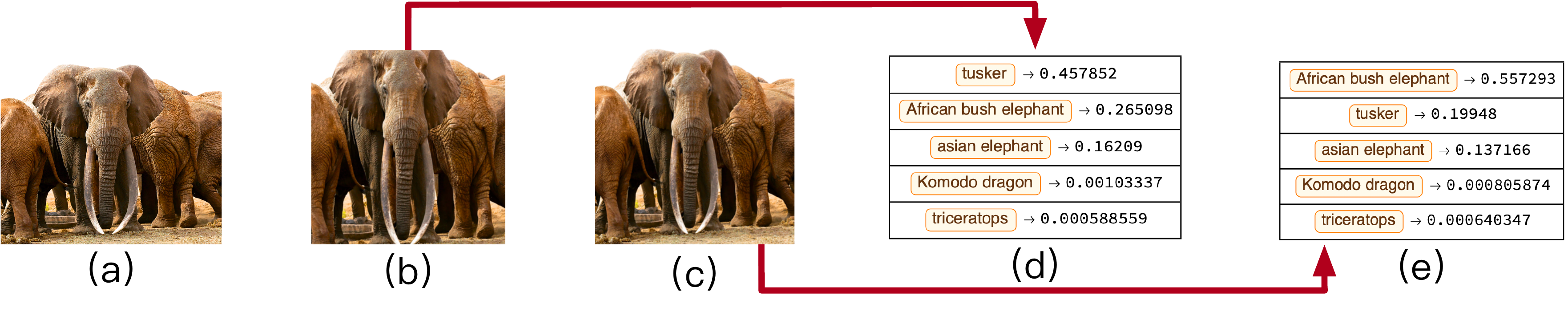}
\vspace{-10pt}
\caption{Differences in the prediction results due to cropping using TensorFlow Inception-v3.}
\label{fig:diff_output_process_pred}
\end{minipage}

\begin{minipage}[t]{\linewidth}
\centering
\includegraphics[width=0.95\textwidth]{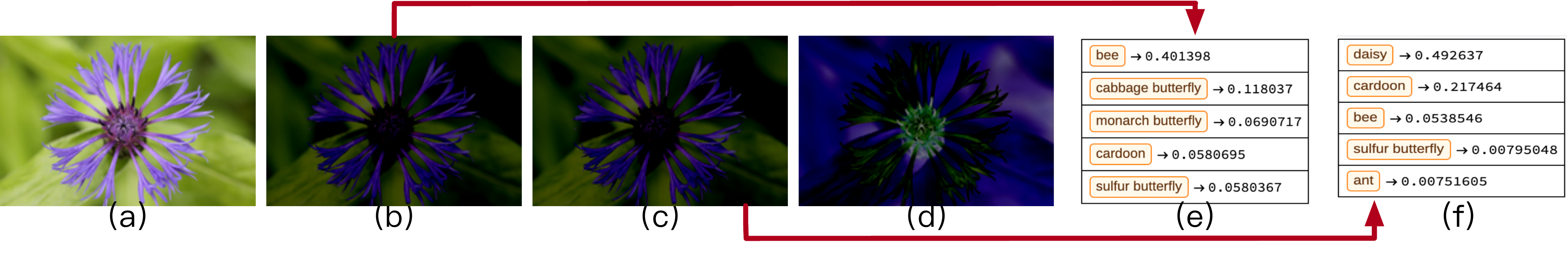}
\vspace{-10pt}
\caption{Differences due to order of operations for data type conversion and normalization using TensorFlow Inception-v3.}
\label{fig:diff_output_process}
\end{minipage}

\begin{minipage}[t]{\linewidth}
\centering
\vspace{-3pt}
\includegraphics[width=\textwidth]{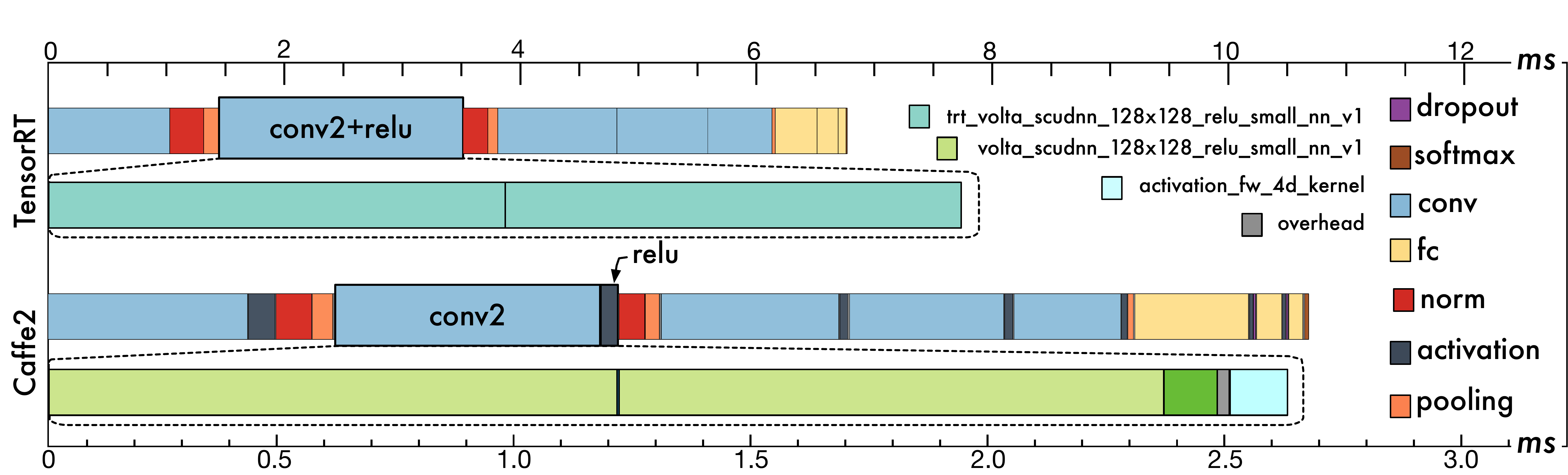}
\caption{Performance of AlexNet running using TensorRT and Caffe2. \carml enables one to understand and debug performance at layer and sub-layer granularity.}
\label{fig:tensorrt_vs_caffe2_framework_compare}
\end{minipage}

\vspace{-25pt}
\end{figure}

\subsection{Effects of Evaluation Pitfalls}\label{sec:pitfalls}

We use Inception-v3~\cite{silberman2016tensorflow,szegedy2016rethinking} to examine the effects of the pitfalls described in Section~\ref{sec:spec}.
Unless otherwise noted, all results use Python $3.6.8$; TensorFlow \texttt{1.13.0-rc2} compiled from source; CUDNN $7.4$; GCC $6.4.0$; Intel Core i7-7820X CPU with Ubuntu $18.04.1$; NVIDIA TITAN V GPU with CUDA Driver $410.72$; and CUDA Runtime $10.0.1$.
When describing the pitfalls, we maintain everything as constant with the exception to the parameters affecting the pitfall.

\subsubsection{Pre/Post-Processing}

Common pre-processing operations (for image classification) are image decoding, cropping, resizing, normalization, and data type conversion.
The model output is a tensor of dimensions $[batch * numClasses]$ which is sorted to get the top $K$ predictions (label with probability).
The effects of the pre-processing pitfalls are:

\textbf{Color Layout} --- 
Models are trained with decoded images that are   in either RGB or BGR  layout.
For legacy reasons, by default OpenCV decodes images in BGR layout and, subsequently, both Caffe and Caffe2 use the BGR layout~\cite{caffebgr}.
Other frameworks (such as TensorFlow or PyTorch) use RGB layout.
Intuitively, objects that are defined by their colors, such as lemons, are likely to be misclassified.
Figure~\ref{fig:lemon} shows the Top 5 inference results for the same image, but with different color layouts.

\textbf{Data Layout} --- The data layout for a two-dimensional image (to be used as a tensor for a model) is represented by: $N$ (number of inputs processed), $C$ (channel), $H$ (number of pixels in vertical dimension), and $W$ (number of pixels in horizontal dimension).
Models are trained with input in either \texttt{NCHW} or \texttt{NHWC} data layout.
Figure~\ref{fig:platapus} shows the Inception-v3 Top1 inference results using different layouts for the same input image, with the model trained with \texttt{NHWC} layout.

\textbf{Image Decoding and Color Conversion} --- It is common to use JPEG as the image data serialization format (with ImageNet being stored as JPEG images).
Model developers use \texttt{opencv.imread}, \texttt{PIL.Image.open}, or \texttt{tf.io.decode\_jpeg} functions to decode the JPEG images.
The underlying JPEG decode method varies across frameworks and systems.
For example, the JPEG decompression method affects the decoded image quality and is system dependent in TensorFlow~\cite{tflibjpeg} (being either \texttt{INTEGER\_FAST} or \texttt{INTEGER\_ACCURATE}~\cite{wallace1992jpeg}).
Even if the underlying decoding method is the same, different libraries may have different implementations of color conversion methods.
For example, we find the YCrCb to RGB color conversion to not be consistent across libraries.
Figure~\ref{fig:pil_vs_opencv_diff} shows the results of decoding an image using Python's PIL and compares it to decoding with OpenCV.
In Figure~\ref{fig:pil_vs_opencv_diff}, edge pixels are not encoded consistently, even though these are critical pixels for models such as object detection.

\textbf{Cropping} --- For image classification,  accuracy is sometimes reported for cropped validation datasets. 
The cropping method and parameter are often overlooked by model evaluators, which results in different accuracy numbers. 
For Inception-v3, for example, the input images are center-cropped with fraction $87.5\%$, and then resized to $299\times299$. 
Figure~\ref{fig:diff_output_process_pred} shows the effect of omitting cropping from pre-processing: (a) is the original image; (b) is the result of center cropping the image with $87.5\%$ and then resizing; (c) is the result of just resizing; (d) and (f) shows the prediction results using processed images from (b) and (c).
Intuitively, cropping differences are more pronounced for input images  where the marginal regions are meaningful (e.g. paintings within frames).

\textbf{Type Conversion and Normalization} ---
After decoding, the image data is in bytes and is converted to FP32 (assuming an FP32 model) before being used as an input to the model. 
Mathematically, float to byte conversion is $float2byte(x) = 255x$, and byte to float conversion is $byte2float(x) = \frac{x}{255.0}$.
Because of programming semantics, however, the executed behavior of byte to float conversion is $byte2float(x) = \floor{\frac{x}{255.0}}$.
As part of the pre-processing, the input may also need to be normalized to have zero mean and unit variance ($\frac{pixel-mean}{stddev}$). 
We find that the order of operations for type conversion and normalization matters.
Figure~\ref{fig:diff_output_process} shows the image processing results using different order of operations for $meanByte = stddevByte = 127.5$ and $meanFloat = stddevFloat = 0.5$ where: (a) is the original image, (b) is the result of reading the image in bytes then normalizing it with the mean and standard deviation in bytes, $byte2float(\frac{imgByte - meanByte}{stddevByte})$, (c) is the result of reading an image in floats then normalizing it with the mean and standard deviation in floats, $\frac{byte2float(imgByte) - meanFloat}{stddevFloat}$, and (d) is the difference between (b) and (c)\footnote{\small To increase the contrast of the differences on paper, we dilate the image (with radius $2$) and rescaled its pixel values to cover the range between $0$ and $1$.}. 
The inference results of Figure~\ref{fig:diff_output_process_pred} (b,c) are shown in Figure~\ref{fig:diff_output_process_pred} (e,f).\abdul{this is not using tensorflow, but mxnet}

\subsubsection{Software Stack}
DL layers across frameworks may have different implementations or dispatch to different library functions.
Figure~\ref{fig:tensorrt_vs_caffe2_framework_compare} shows 
AlexNet's latency across different frameworks with Caffe2 performing the inference in $10.61ms$ whereas TensorRT performs it in $6.75ms$.  
\carml is able to help users figure out the source of the latency drop.
Take the \textit{conv2} and the following \textit{relu} layers for example.
In TensorRT, these two layers are fused and are mapped to two \texttt{trt\_volta\_scudnn\_128x128\_relu\_small\_nn\_v1} kernels that take $1.95ms$. 
In Caffe2, however, the layers are not fused and take $2.63ms$. 
Understanding the performance across the HW/SW stack is key to comparing hardware or software performance.%

\begin{wrapfigure}{hr}{0.43\textwidth}
\vspace{-50pt}
  \centering
  \tikzsetnextfilename{latex_figures/fig_language_cpu}%
  \tikzpicturedependsonfile{latex_figures/fig_language_cpu.tex}%
  \begin{subfigure}[b]{0.5\textwidth}
\centering
\begin{tikzpicture}
\tikzsetfigurename{language_cpu_inceptionv3_tensorflow}
\begin{axis}[
    ymajorgrids=true,
    grid style=dashed,
	ybar interval=0.7,
	xlabel=(b) Normalized CPU Latency.,
	font=\sffamily,
	enlargelimits=0.05,
	tick label style={font=\footnotesize},
	label style={font=\small},
    legend columns=5,
	width=0.8\linewidth,
    height=0.2\linewidth,
    axis x line*=bottom,
    axis y line=left,
    legend style={
    at={(0,0)},anchor=north west,at={(axis description cs:0.16,1.3)},
    font=\sffamily\footnotesize,fill=none,draw=none},
	cycle list name=Dark2-8,
	every axis plot/.append style={fill,draw=none,no markers},
	symbolic x coords = {1,2,4,8,16,32,999},
    xtick=data,
    bar shift=0pt,
    bar width = 10pt,
    ymin=0,
    legend image code/.code={%
        \draw[#1, draw=none] (0cm,-0.1cm) rectangle (0.2cm,0.1cm);
    },  
]
	\addplot[plotcolor1] table [x index=0,y index=4] {language_cpu_inceptionv3_tensorflow.dat};
	\addlegendentry{C++}

	\addplot[plotcolor3] table [x index=0,y index=5] {language_cpu_inceptionv3_tensorflow.dat};
	\addlegendentry{NumPy} 
	
	\addplot[plotcolor4] table [x index=0,y index=6] {language_cpu_inceptionv3_tensorflow.dat};
	\addlegendentry{Python}

\end{axis}
\end{tikzpicture} 
\vspace{-0.1in}
\label{fig:eval_cpu_language}
\end{subfigure}%

\\
  \centering
  \tikzsetnextfilename{latex_figures/fig_language_gpu}%
  \tikzpicturedependsonfile{latex_figures/fig_language_gpu.tex}%
  \begin{subfigure}[b]{0.5\textwidth}
\centering
\begin{tikzpicture}
\tikzsetfigurename{language_gpu_inceptionv3_tensorflow}
\begin{axis}[
    ymajorgrids=true,
    grid style=dashed,
	ybar interval=0.7,
	xlabel=(b) Normalized GPU Latency.,
	font=\sffamily,
	enlargelimits=0.05,
	tick label style={font=\footnotesize},
	label style={font=\small},
	width=0.8\linewidth,
    height=0.2\linewidth,
    axis x line*=bottom,
    axis y line=left,
	cycle list name=Dark2-8,
	every axis plot/.append style={fill,draw=none,no markers},
	symbolic x coords = {1,2,4,8,16,32,999},
    xtick=data,
    bar shift=0pt,
    bar width = 10pt,
    ymin=0,
    legend columns=5,
    legend style={
    at={(0,0)},anchor=north west,at={(axis description cs:0,1.25)},
    font=\sffamily\footnotesize,fill=none,draw=none},
    legend image code/.code={%
        \draw[#1, draw=none] (0cm,-0.1cm) rectangle (0.2cm,0.1cm);
    },  
]
	\addplot[plotcolor1] table [x index=0,y index=4] {language_gpu_inceptionv3_tensorflow.dat};

	\addplot[plotcolor3] table [x index=0,y index=5] {language_gpu_inceptionv3_tensorflow.dat};
	
	\addplot[plotcolor4] table [x index=0,y index=6] {language_gpu_inceptionv3_tensorflow.dat};

\end{axis}
\end{tikzpicture} 
\label{fig:eval_gpu_language}
\end{subfigure}%

\vspace{-0.3in}
    \caption{\texttt{tf.Session.Run} execution time (normalized to C/C++) vs. batch size for Inception-v3 inference on CPU and GPU using TensorFlow with C++, Python using NumPy arrays, and Python using native lists.}
\vspace{-10pt}
\label{fig:c_vs_python}
\end{wrapfigure}
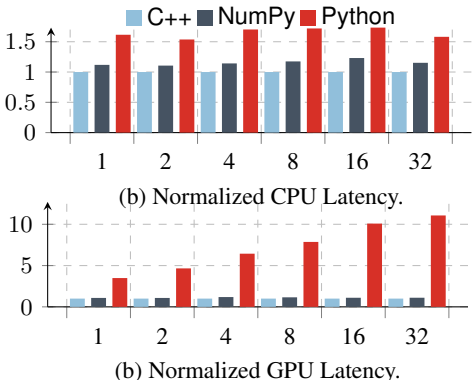

\subsubsection{Programming Language}

Figure~\ref{fig:c_vs_python} shows the normalized inference latency across language environments on GPU and CPU for different batch sizes.
The C/C++ GPU implementation achieves a $5-24\times$ speedup over the CPU only version.
On CPU, Python is $64\%$ and NumPy is $15\%$ slower than C/C++; whereas on GPU Python is $3-11\times$ and NumPy is $10\%$ slower than C/C++.
For Python, the overhead is proportional to the input size and is due to TensorFlow internally having to unbox the Python linked list objects and create a numeric buffer that can be used by the C/C++ code.
The unboxing is not needed for NumPy since TensorFlow can use NumPy's internal numeric buffer directly. 
One should use C++ for latency sensitive production code or for bare-metal benchmarking.

\subsection{Case Study 1: Sharing Models}\label{sec:casestudy_share_model}

During the course of adding models to \carml, we had to verify the accuracy of hundreds of models against  model owners’ claims.
The experience of this verification process motivated our design of \carml and re-enforced our belief that a better way of sharing ML models is needed, and that we are solving a real problem.
To verify the ease of writing an evaluation specification,
we asked our colleagues who develop ML/DL models to integrate their models within \carml.
Our colleagues usually spent less than $10$ minutes to prepare a model manifest and deploy the model using \carml’s web UI.
We were then able to fully reproduce their reported accuracy without their  consultation. 
This shows that \carml provides an intuitive system for model owners to share their models, without requiring too much time on their part.

\begin{table}
    \vspace{-10pt}
    \centering
    \resizebox{\textwidth}{!}{%
    
\begin{tabular}{lcccccccccc} \toprule
\centering%

 & \multicolumn{2}{c}{{{\bfseries Reported}}}
 & \multicolumn{2}{c}{{{\bfseries Measured w/\carml}}}
 & \multicolumn{2}{c}{{{\bfseries Color Layout Pitfall}}}
 & \multicolumn{2}{c}{{{\bfseries Cropping  Pitfall }}}
 & \multicolumn{2}{c}{{{\bfseries Type Conversion Pitfall}}} \\

\cmidrule[0.4pt](lr{0.125em}){2-3}%
\cmidrule[0.4pt](lr{0.125em}){4-5}%
\cmidrule[0.4pt](lr{0.125em}){6-7}%
\cmidrule[0.4pt](lr{0.125em}){8-9}%
\cmidrule[0.4pt](lr{0.125em}){10-11}%

\centering%
     \textbf{Model Name} & \textbf{Top1} & \textbf{Top5} & \textbf{Top1} & \textbf{Top5}  & \textbf{Top1} & \textbf{Top5}   & \textbf{Top1} & \textbf{Top5}   & \textbf{Top1} & \textbf{Top5}    \\ \midrule

Inception-V3~\cite{szegedy2016rethinking}  &  $78.77\%$  &  $94.39\%$  &  $78.41\%$   &  $94.07\%$   &  $67.44\%$   &  $88.44\%$  &  $78.27\%$   &  $94.24\%$   &  $78.41\%$   &  $94.08\%$  \\
MobileNet1.0~\cite{howard2017mobilenets}   &  $73.28\%$  &  $91.30\%$  &  $73.27\%$   &  $91.30\%$   &  $59.22\%$   &  $82.95\%$  &  $71.26\%$   &  $90.17\%$   &  $73.27\%$   &  $91.29\%$  \\
ResNet50-V1~\cite{he2016deep}  &  $77.36\%$  &  $93.57\%$  &  $77.38\%$   &  $93.58\%$   &  $63.21\%$   &  $85.65\%$  &  $75.87\%$   &  $92.82\%$   &  $77.40\%$   &  $93.56\%$  \\
ResNet50-V2~\cite{he2016identity}  &  $77.11\%$  &  $93.43\%$  &   $77.15\%$   &  $93.43\%$   &  $63.35\%$   &  $85.95\%$  &  $75.71\%$   &  $92.72\%$   &  $77.13\%$   &  $93.42\%$  \\
VGG16~\cite{simonyan2014very}  &  $73.23\%$  &  $91.31\%$  &  $73.23\%$   &  $91.31\%$   &  $59.17\%$   &  $82.77\%$  &  $71.71\%$   &  $90.61\%$   &  $73.24\%$   &  $91.33\%$  \\
VGG19~\cite{simonyan2014very}  &  $74.11\%$   &  $91.35\%$  &  $74.15\%$   &  $91.77\%$   &  $60.41\%$   &  $83.57\%$   &  $72.66\%$   &  $90.99\%$   &  $74.14\%$   &  $91.75\%$  \\
\bottomrule
\end{tabular}%
	}
    \caption{
        The effects of the pitfalls on the Top 1 and Top 5 accuracy for heavily cited DL models.
    }
  \setlength{\belowcaptionskip}{-5pt}
    \label{tab:models}
    \vspace{-25pt}
\end{table}

\subsection{Case Study 2: Evaluating Models}\label{sec:casestudy_eval_model}

To show \carml's evaluation capability and the effects of the pitfalls on accuracy, we install it on multiple Amazon P3~\cite{awsp3} instances.
We then created multiple variants of the manifest file that trigger the pitfalls and evaluate the Top 1 and Top 5 accuracy for the models across the systems.
Table~\ref{tab:models} shows that \carml reproduces reported results within $0\%-5.5\%$ for Top 1 accuracy, and $0\%-3.1\%$ for Top 5 accuracy.
Absent a specification or detailed description, it is hard for \carml to fully reproduce the reported results and/or identify the cause of the  accuracy discrepancy.
It is also easy for model evaluators to fall into one or more of the pitfalls outlined in this paper.
Table~\ref{tab:models} shows the accuracy errors are close for some pitfalls, but, in general, they are significantly higher than \carml's~\footnote{\small 
We omit from Table 1 the data layout pitfall results, which, as expected, has very low accuracy.}.
For example, failure to center-crop the input results in $1.45\%-7.5\%$ and Top 1 accuracy difference, and $0.36\%-4.22\%$ Top 5 accuracy difference.

\vspace{-2pt}

\section{Related Work}\label{sec:related}

Prior works are mainly  repositories of curated models (called model zoos). 
These framework specific model  zoos~\cite{caffe2zoo,caffezoo,gluoncv,marcel2010torchvision,onnxzoo, tensorflowhub} are used for testing or demonstrating the kinds of models a framework supports.
There are also repositories of model zoos~\cite{modelhub, modelzoo} or public hubs linking ML/DL papers with their corresponding code~\cite{paperswithcode}.
In~\cite{miao2017modelhub,tsay2018runway,vartak2016m}, the authors manage ML/DL artifacts (models, data, or experiments) and their provenance.
~\cite{church2017emerging,mannarswamy2018evolving,sethi2018dlpaper2code} discuss methodologies to simplify evaluation of ML/DL research.
However, none of the previous work try to standardize ML/DL model evaluations, or provide a coherent way of specifying, provisioning, and running model evaluations.
We’re the first to propose such a system design.
To facilitate reproducible research and simplify paper evaluation, conferences such as NeurIPS, ICLR, and AAAI develop guidelines or checklists for paper authors to follow.
These conferences encourage the model owners to publish code, as well as to specify the system requirements to run it.
However, this has some disadvantages:
\circledwhite{1} A self-imposed guideline is difficult to maintain.
If it places a lot of burden on the model owner, then no one will use it.
But, if it leaves too much at their discretion, then there will be variability in quality and scope of the evaluation procedure description.
\circledwhite{2} 
Defining the execution parameters as text is a lot of effort, but, due to the lack of standardization, defining them in code leads to a non-uniform evaluation procedure --- making it difficult to perform fair comparisons across models and HW/SW stacks.
\circledwhite{3} As we showed in this paper, there are subtle evaluation parameters that need to be carefully specified to guarantee that model evaluation results match with the owner's claims.

\vspace{-2pt}

\section{Conclusion and Future Work}\label{sec:conclusion}

ML/DL is fast advancing with new models, software stacks, and hardware being introduced daily.
In the future, the problems outlined are only going to be compounded with the introduction of more diverse sets of frameworks, models, and HW/SW stacks.
We therefore see that both a specification along with a fast and consistent way to evaluate models (without putting too much burden on model owners) are key to manage both the complexity and the fast pace of  ML/DL.
This paper described limitations of current practice of sharing models.
It then proposes a solution --- \carml, which is a specification, runtime, and data aggregation design that is both scalable, extensible, and easy-to-use.
Through \carml's evaluation specification, model owners can define parameters needed to evaluate their models and share their models, while guaranteeing repeatable evaluation.
\carml also helps the general model evaluators in provisioning the system stack to evaluate the model using the specification provided by the model owner.
Future work would leverage data gathered from \carml to automate algorithm and hardware suggestion (based on user input), perform  intelligent resource scheduling for ML/DL application workflows, and address any newly discovered pitfalls and vulnerabilities.

\small
\bibliographystyle{acm}
\bibliography{main} 

\begin{thebibliography}{10}

\bibitem{awsp3}
{Amazon EC2 P3 Instances}.
\newblock \url{https://aws.amazon.com/ec2/instance-types/p3/}, 2019.
\newblock Accessed: 2019-05-22.

\bibitem{caffe2zoo}
Caffe2 model zoo.
\newblock \url{https://caffe2.ai/docs/zoo.html}, 2019.
\newblock Accessed: 2019-05-22.

\bibitem{caffebgr}
{Image Pre-Processing}.
\newblock \url{https://caffe2.ai/docs/tutorial-image-pre-processing.html},
  2019.
\newblock Accessed: 2019-05-22.

\bibitem{caffezoo}
{Caffe Model Zoo}.
\newblock \url{https://caffe.berkeleyvision.org/model_zoo.html}, 2019.
\newblock Accessed: 2019-05-22.

\bibitem{church2017emerging}
{\sc Church, K.~W.}
\newblock {Emerging trends: I did it, I did it, I did it, but...}
\newblock {\em Natural Language Engineering 23}, 3 (2017), 473--480.

\bibitem{gluoncv}
{GluonCV}.
\newblock \url{https://gluon-cv.mxnet.io/}, 2019.
\newblock Accessed: 2019-05-22.

\bibitem{he2016deep}
{\sc He, K., Zhang, X., Ren, S., and Sun, J.}
\newblock Deep residual learning for image recognition.
\newblock In {\em Proceedings of the IEEE conference on computer vision and
  pattern recognition\/} (2016), pp.~770--778.

\bibitem{he2016identity}
{\sc He, K., Zhang, X., Ren, S., and Sun, J.}
\newblock Identity mappings in deep residual networks.
\newblock In {\em European conference on computer vision\/} (2016), Springer,
  pp.~630--645.

\bibitem{howard2017mobilenets}
{\sc Howard, A.~G., Zhu, M., Chen, B., Kalenichenko, D., Wang, W., Weyand, T.,
  Andreetto, M., and Adam, H.}
\newblock Mobilenets: Efficient convolutional neural networks for mobile vision
  applications.
\newblock {\em arXiv preprint arXiv:1704.04861\/} (2017).

\bibitem{mannarswamy2018evolving}
{\sc Mannarswamy, S., and Roy, S.}
\newblock {Evolving AI from Research to Real Life-Some Challenges and
  Suggestions.}
\newblock In {\em IJCAI\/} (2018), pp.~5172--5179.

\bibitem{marcel2010torchvision}
{\sc Marcel, S., and Rodriguez, Y.}
\newblock Torchvision the machine-vision package of torch.
\newblock In {\em Proceedings of the 18th ACM international conference on
  Multimedia\/} (2010), ACM, pp.~1485--1488.

\bibitem{miao2017modelhub}
{\sc Miao, H., Li, A., Davis, L.~S., and Deshpande, A.}
\newblock Modelhub: Deep learning lifecycle management.
\newblock In {\em 2017 IEEE 33rd International Conference on Data Engineering
  (ICDE)\/} (2017), IEEE, pp.~1393--1394.

\bibitem{modelhub}
{ Modelhub}.
\newblock \url{http://modelhub.ai}, 2019.
\newblock Accessed: 2019-05-22.

\bibitem{modelzoo}
{ModelZoo}.
\newblock \url{https://modelzoo.co}, 2019.
\newblock Accessed: 2019-05-22.

\bibitem{onnxzoo}
{ONNX Model Zoo}.
\newblock \url{https://github.com/onnx/models}, 2019.
\newblock Accessed: 2019-05-22.

\bibitem{opentracing}
{OpenTracing}: Cloud native computing foundation.
\newblock \url{http://opentracing.io}, 2019.
\newblock Accessed: 2019-05-22.

\bibitem{paperswithcode}
{Papers with Code}.
\newblock \url{https://paperswithcode.com}, 2019.
\newblock Accessed: 2019-05-22.

\bibitem{semanticver}
{\sc Preston-Werner, T.}
\newblock Semantic versioning 2.0.0.
\newblock \url{https://www.semver.org}, 2019.

\bibitem{sethi2018dlpaper2code}
{\sc Sethi, A., Sankaran, A., Panwar, N., Khare, S., and Mani, S.}
\newblock Dlpaper2code: Auto-generation of code from deep learning research
  papers.
\newblock In {\em Thirty-Second AAAI Conference on Artificial Intelligence\/}
  (2018).

\bibitem{silberman2016tensorflow}
{\sc Silberman, N., and Guadarrama, S.}
\newblock Tensorflow-slim image classification model library.
\newblock {\em URL:
  https://github.com/tensorflow/models/tree/master/research/slim\/} (2018).

\bibitem{simonyan2014very}
{\sc Simonyan, K., and Zisserman, A.}
\newblock Very deep convolutional networks for large-scale image recognition.
\newblock {\em arXiv preprint arXiv:1409.1556\/} (2014).

\bibitem{szegedy2016rethinking}
{\sc Szegedy, C., Vanhoucke, V., Ioffe, S., Shlens, J., and Wojna, Z.}
\newblock Rethinking the inception architecture for computer vision.
\newblock In {\em Proceedings of the IEEE conference on computer vision and
  pattern recognition\/} (2016), pp.~2818--2826.

\bibitem{tflibjpeg}
{tf.io.decode\_jpeg}.
\newblock \url{https://www.tensorflow.org/api_docs/python/tf/io/decode_jpeg},
  2019.
\newblock Accessed: 2019-05-22.

\bibitem{tensorflowhub}
{TensorFlow Hub}.
\newblock \url{https://www.tensorflow.org/hub}, 2019.
\newblock Accessed: 2019-05-22.

\bibitem{tsay2018runway}
{\sc Tsay, J., Mummert, T., Bobroff, N., Braz, A., Westerink, P., and Hirzel,
  M.}
\newblock Runway: machine learning model experiment management tool.
\newblock {\em SysML\/} (2018).

\bibitem{vartak2016m}
{\sc Vartak, M., Subramanyam, H., Lee, W.-E., Viswanathan, S., Husnoo, S.,
  Madden, S., and Zaharia, M.}
\newblock Modeldb: a system for machine learning model management.
\newblock In {\em Proceedings of the Workshop on Human-In-the-Loop Data
  Analytics\/} (2016), ACM, p.~14.

\bibitem{wallace1992jpeg}
{\sc Wallace, G.~K.}
\newblock {The JPEG still picture compression standard}.
\newblock {\em IEEE transactions on consumer electronics 38}, 1 (1992),
  xviii--xxxiv.

\end{thebibliography}

\clearpage
\newpage

\end{document}